\documentclass[letterpaper]{article} 
\usepackage{aaai2026}  
\usepackage{times}  
\usepackage{helvet}  
\usepackage{courier}  
\usepackage[hyphens]{url}  
\usepackage{graphicx} 
\usepackage{natbib}  
\usepackage{caption} 
\usepackage{algorithm}
\usepackage{algorithmic}

\usepackage{newfloat}
\usepackage{listings}
\DeclareCaptionStyle{ruled}{labelfont=normalfont,labelsep=colon,strut=off} 
\floatstyle{ruled}
\newfloat{listing}{tb}{lst}{}
\floatname{listing}{Listing}
\usepackage{tabularx}
\usepackage{pifont}
\usepackage{amsmath}
\usepackage{xspace}
\usepackage{multirow}
\usepackage{makecell}
\usepackage{amsfonts}
\usepackage{booktabs}

\newcommand{\ourarch}[1]{\textbf{\textit{CLEAR-Mamba}}}

\newcommand{\model}[1]{\textbf{\textit{#1}}}

\title{\ourarch{}: Towards Accurate, Adaptive and Trustworthy Multi-Sequence Ophthalmic Angiography Classification}

\author{
  Zhuonan Wang\textsuperscript{\rm 1}\equalcontrib,
  Wenjie Yan\textsuperscript{\rm 1}\equalcontrib,
  Wenqiao Zhang\textsuperscript{\rm 1},
  Xiaohui Song\textsuperscript{\rm 1,2},\\
  Jian Ma\textsuperscript{\rm 1,2},
  Ke Yao\textsuperscript{\rm 1,2},
  Yibo Yu\textsuperscript{\rm 1,2}\thanks{Corresponding author.},
  Beng Chin Ooi\textsuperscript{\rm 1}
}
\affiliations{
    \textsuperscript{\rm 1}Zhejiang University, \textsuperscript{\rm 2}Eye Center of the Second Affiliated Hospital\\
    \{wnzz, 22551068, wenqiaozhang, 2317050, jian\_ma, xlren, yuyibo, ooibc\}@zju.edu.cn\\
}

\usepackage{bibentry}

\nocopyright

\begin{document}

\makeatletter

{
\renewcommand\@makefntext[1]{%
  \noindent\@makefnmark\hspace{0.6em}#1%
}
\maketitle
}
\makeatother

\begin{abstract}
Medical image classification is a core task in computer-aided diagnosis (CAD), playing a pivotal role in early disease detection, treatment planning, and patient prognosis assessment.
In ophthalmic practice, fluorescein fundus angiography (FFA) and indocyanine green angiography (ICGA) provide hemodynamic and lesion-structural information that conventional fundus photography cannot capture. 
However, due to the single-modality nature, subtle lesion patterns, and significant inter-device variability, existing methods still face limitations in generalization and high-confidence prediction.
To address these challenges, we propose \ourarch{}, an enhanced framework built upon MedMamba with optimizations in both architecture and training strategy. 
Architecturally, we introduce \textbf{HaC}, a hypernetwork-based adaptive conditioning layer that dynamically generates parameters according to input feature distributions, thereby improving cross-domain adaptability.
From a training perspective, we develop \textbf{RaP}, a reliability-aware prediction scheme built upon evidential uncertainty learning, which encourages the model to emphasize low-confidence samples and improves overall stability and reliability.
We further construct a large-scale ophthalmic angiography dataset covering both FFA and ICGA modalities, comprising multiple retinal disease categories for model training and evaluation.
Experimental results demonstrate that \ourarch{} consistently outperforms multiple baseline models, including the original MedMamba, across various metrics—showing particular advantages in multi-disease classification and reliability-aware prediction.
This study provides an effective solution that balances generalizability and reliability for modality-specific medical image classification tasks.
Our project can be accessed at \url{https://github.com/ZJU4HealthCare/CLEAR-Mamba}.
\end{abstract}

\section{Introduction}
\label{intro}
Recent advances in multimodal foundation models, driven by representation learning and large-scale pretraining, have enabled unified perception and reasoning across vision, language, and domain knowledge, demonstrating strong generalization across diverse applications.
~\cite{lin2025healthgpt,xie2025heartcare,li2025eyecaregpt,zhang2022boostmis,yuan2025pixelrefer,yuan2025videorefer}. 
Medical image classification is a fundamental task in computer-aided diagnosis (CAD), aiming to automatically recognize disease types or pathological conditions from imaging data. With recent advances in deep learning, it has shown strong potential across diverse clinical scenarios~\cite{dao2025recent,tang2022fusionm4net}.
Among imaging modalities, ophthalmic angiography is particularly important. FFA and ICGA provide dynamic cues of retinal hemodynamics and choroidal structures, supporting the diagnosis of diseases such as AMD~\cite{wong2014global}, DR~\cite{sivaprasad2012prevalence}, and glaucoma~\cite{thylefors1994global}. Compared with fundus photography, angiography better reveals subtle vascular abnormalities, offering richer evidence for diagnosis and treatment planning~\cite{li2022comparison,mahendradas2021role,invernizzi2023experts}.

The temporal nature of angiography offers unique opportunities for automated classification. However, most existing studies mainly focus on multimodal fusion~\cite{luo2025survey,yuan2025videorefer,yuan2025pixelrefer}. Typical CNN/ViT-based models integrate features from multiple modalities (e.g., CFP, OCT, OCTA, and FFA) to improve performance on AMD, glaucoma, and DR-related tasks. For instance, \cite{jin2022multimodal} fuses OCT and OCTA features to assess CNV activity, while MMRAF~\cite{zhou2023representation} enhances glaucoma recognition via contrastive alignment, multi-instance representation, and hierarchical attention fusion. In addition, \cite{hervella2022multimodal} proposes cross-modal pretraining on unlabeled retinal image pairs to improve generalization.

While multimodal methods exploit complementary cues across modalities, they largely emphasize inter-modality integration. In contrast, a single FFA/ICGA exam naturally forms a sequential series of frames that captures hemodynamics and lesion evolution. Treating angiography as static images can miss temporal signatures (e.g., early filling to late leakage), thus underutilizing the diagnostic value of single-modality angiography.
Meanwhile, mainstream architectures have inherent drawbacks for temporal imaging. CNNs struggle to capture long-range temporal dependencies due to limited receptive fields~\cite{liu2024multi}, whereas ViTs model global context at the cost of heavy computation and memory~\cite{li2023transforming}, limiting real-time and resource-efficient deployment. Together with the underuse of angiography’s temporal dynamics, these issues remain a key bottleneck for clinically practical angiographic classification.

Beyond temporal modeling, medical AI systems also face two key bottlenecks: limited predictive reliability and weak generalizability~\cite{hasani2022trustworthy,zhang2024revisiting}. For reliability, softmax scores are often misused as confidence, leading to overconfident predictions on noisy or out-of-distribution inputs~\cite{guo2017calibration} and failing to disentangle epistemic and aleatoric uncertainty~\cite{kendall2017uncertainties}. In high-stakes settings, this can produce confident yet unreliable decisions, motivating standardized and quantifiable confidence estimation, e.g., calibration and uncertainty-aware deferral to human review when confidence is low.
Equally important is generalizability. Many studies target single-disease classification (e.g., AMD, glaucoma, or DR), which often performs well in-domain but degrades in multi-disease or more complex clinical settings~\cite{yang2022machine,ong2024shortcut}. 
Recent efforts have explored parameter generation mechanisms to improve model adaptability without explicit fine-tuning, such as dynamically generating model parameters conditioned on domain or input characteristics~\cite{lv2023duet}. However, these approaches are primarily designed for device-level domain shifts and do not account for the temporal progression patterns or uncertainty-aware decision requirements inherent to ophthalmic angiography.

Taken together, conventional deep learning architectures and modeling paradigms remain inadequate for the comprehensive demands of ophthalmic angiography classification. Future methods must achieve breakthroughs in three core areas—making more effective use of temporal information, providing reliable confidence estimation, and enhancing adaptability across diverse clinical scenarios—before they can be truly translated into clinical practice.

To address these limitations, we propose \ourarch{} for single-modality, multi-sequence angiography, aiming to improve temporal modeling, case adaptability, and prediction reliability. We curate a 43-disease dataset with complete FFA/ICGA sequences and adopt MedMamba as the backbone to efficiently capture long-range dynamics. We further introduce a HyperNetwork for lightweight, case-conditioned adaptation~\cite{zhang2024hyperllava} and an evidential learning head to output calibrated predictions with uncertainty, enabling risk-aware deferral in ambiguous cases.

In summary, \ourarch{} integrates large-scale multi-sequence angiography data, the efficient modeling capacity of the Mamba backbone, the adaptive representation of HyperNetworks, and the uncertainty quantification of evidential learning into a unified framework, offering a more robust and clinically oriented solution for angiographic image classification.

The main contributions are summarized as follows:
\begin{itemize}
    \item \textbf{Method:} We propose \ourarch{}, built on MedMamba for efficient temporal modeling, augmented with a HyperNetwork for lightweight case-wise adaptation and evidential learning for calibrated confidence and uncertainty-aware predictions.
    
    \item \textbf{Dataset:} We curate a large-scale single-modality, multi-sequence FFA/ICGA dataset covering 43 diseases with complete temporal sequences to capture hemodynamics and lesion progression.
    
    \item \textbf{Results:} Extensive experiments on our dataset and public benchmarks show consistent improvements over strong CNN/ViT/Mamba baselines (including MedMamba) in accuracy and reliability, demonstrating better temporal utilization, generalization, and confidence modeling.
\end{itemize}
\section{Related Work}
\label{related work}

\subsection{Medical Image Classification}
Medical image classification (MIC) is a core task in computer-aided diagnosis, with deep learning substantially improving diagnostic accuracy and efficiency~\cite{dao2025recent}. Early studies predominantly relied on CNNs~\cite{lecun2002gradient}, while later works addressed robustness issues such as label noise~\cite{liao2025unleashing}.
The introduction of Vision Transformers (ViTs)~\cite{dosovitskiy2020image} further advanced MIC by enhancing global representation modeling, leading to improved performance in complex modalities such as CT and MRI~\cite{manzari2023medvit}. More recently, alternative backbones have been explored: MedMamba~\cite{yue2024medmamba} leverages structured state space models~\cite{gu2023modeling} to better handle long sequences. Despite these advances, most existing MIC methods remain single-modality and do not fully exploit multi-temporal characteristics, particularly in ophthalmic angiography. Meanwhile, growing clinical demands have motivated efforts toward better explainability and generalization~\cite{rudin2019stop,van2022explainable}.

\subsection{Ophthalmic Diagnostic Models}
Recent ophthalmic diagnostic models increasingly exploit multimodal data (e.g., CFP, OCT, and angiography) to improve classification accuracy and generalization~\cite{luo2025survey}. Early studies mainly focused on single-disease scenarios, such as AMD~\cite{chen2021multimodal}, glaucoma~\cite{mehta2021automated}, and diabetic retinopathy~\cite{hua2020convolutional}, demonstrating that cross-modal attention and feature fusion can enhance disease-specific recognition. With growing demands for broader applicability, multi-disease classification has attracted increasing attention. Representative methods such as MSAN~\cite{he2021multi} employ region-guided and multi-scale attention to improve multimodal feature extraction, while EyeMoSt+~\cite{zou2024confidence} further incorporates uncertainty modeling to enhance robustness under noisy conditions.
Despite these advances, multimodal data are often unavailable in routine clinical practice, making single-modality settings more common. This highlights the importance of developing robust and generalizable single-modality models that can operate reliably under practical clinical constraints.

\subsection{Ophthalmic Imaging Datasets}
The development of ophthalmic AI models relies on high-quality and diverse datasets across modalities and disease categories~\cite{luo2025survey}. Early datasets mostly focused on single diseases, such as AREDS~\cite{study1999age} for AMD progression and DeepDRiD~\cite{liu2022deepdrid} for DR grading, with additional resources supporting glaucoma-related tasks.
Later efforts expanded to limited multi-disease settings (e.g., Rabbani-I~\cite{rasti2017macular} and Messidor~\cite{abramoff2013automated}), and more recent datasets better reflect clinical diversity, including large-scale multi-pathology OCT collections~\cite{kermany2018identifying} and multimodal resources such as BioMISA~\cite{hassan2018biomisa} and ROSE~\cite{ma2020rose}.
However, many datasets still suffer from class imbalance, annotation variability, and limited temporal coverage, which hinders robust and generalizable deployment in real-world clinics.
\section{Data Collection and Processing}
\label{dataset}

The dataset was collected from routine clinical case reports at a partner hospital, stored as PDFs containing multiple high-resolution angiography images with accompanying clinical text. It includes both structured attributes (e.g., demographics, exam metadata, modality, diagnosis) and unstructured content (images and narratives), covering FFA and ICGA across diverse retinal and choroidal disorders. As real-world data, it exhibits natural variability in acquisition quality.

\begin{figure}[t]
    \centering
    \includegraphics[width=1\linewidth]{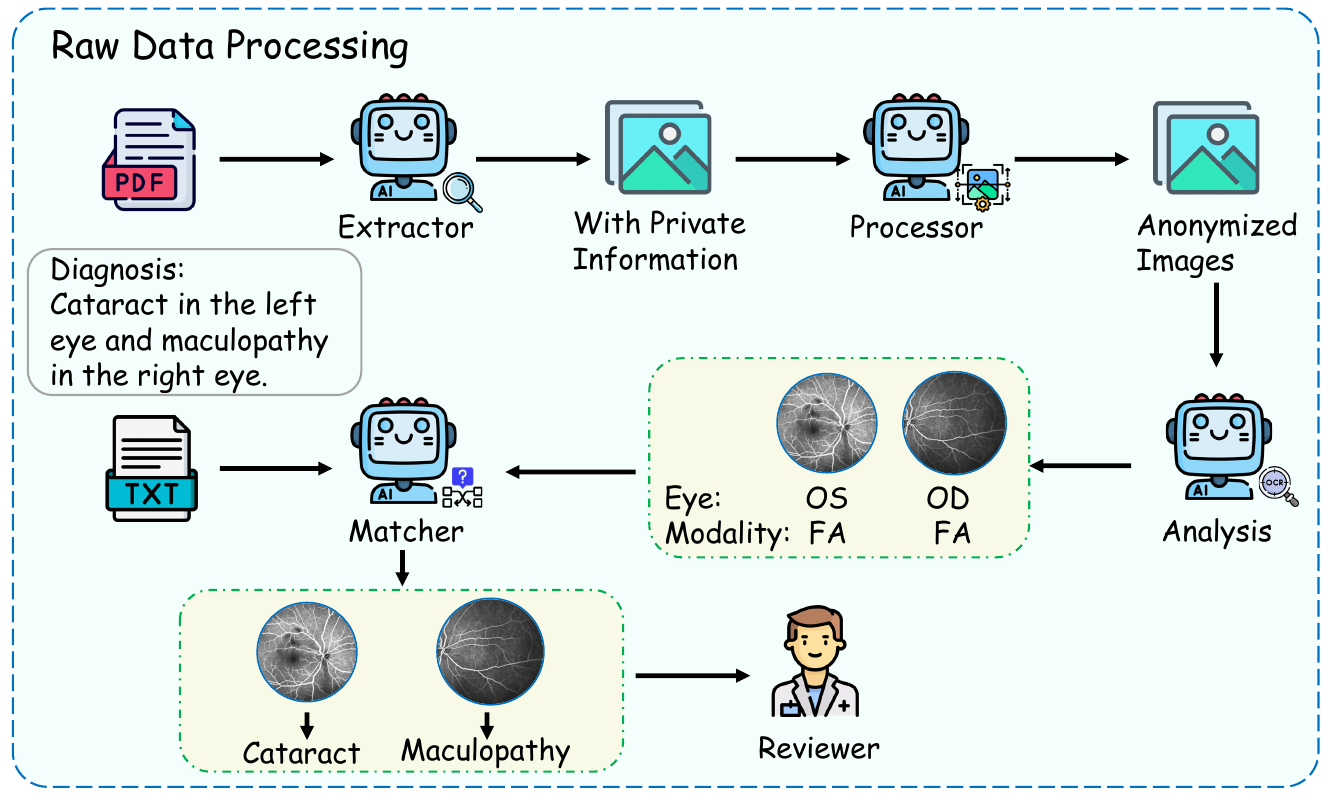}
    \caption{Automated pipeline for extracting, anonymizing, and aligning medical image–text data from raw PDF reports.}
    \label{fig:process}
    \vspace{-3mm}
\end{figure}

Directly using the raw reports for training is challenging: images and text must be extracted and aligned from PDFs, privacy-sensitive identifiers must be removed, and label noise can arise from eye-level annotations within binocular images. Modality and laterality cues are also scattered across the report and require accurate parsing for correct image--label matching. To address this, we build an automated multi-agent pipeline (Fig.~\ref{fig:process}) to convert raw PDFs into a cleaned, structured dataset via extraction, anonymization, alignment, and mismatch filtering/re-annotation.

\subsection{Multi-Agent Data Engine}
We developed an multi-agent data processing engine that converts raw clinical PDF reports into high-quality structured image data. It performs document parsing, text extraction, image anonymization, lesion/label matching, and manual quality control in an end-to-end pipeline, with specialized agents collaborating to improve both efficiency and accuracy.

\subsubsection{Extractor for Image Conversion}
Ophthalmic images are typically embedded in PDF examination reports, which are not directly usable for vision models. This module scans PDFs page by page, detects image regions via layout analysis, and extracts them automatically. The images are then rendered as high-fidelity JPGs while preserving resolution and visual details, ensuring fine-grained lesion cues (e.g., edges and textures) are retained for downstream analysis.

\subsubsection{Analyzer for Description Extraction}
To support lesion screening and label alignment, this module extracts key metadata from reports using OCR and NLP, including laterality (OS/OD) and modality (FFA/ICGA). It handles abbreviations and mixed Chinese--English writing via a standardized terminology dictionary and context-aware semantic matching. The parsed metadata are then bound to the corresponding images for downstream matching and classification.

\subsubsection{Processor for Image Anonymization}
Ophthalmic images may contain identifiers (e.g., names, case IDs, dates) in edge overlays. This module automatically detects and masks/crops these regions using rule-based templates and detection models. To preserve diagnostic content, it first localizes retinal boundaries and avoids removing clinically relevant areas, while also cleaning residual device UI/backgrounds to produce privacy-safe images with retained diagnostic value.

\subsubsection{Matcher for Binocular Lesion Screening}
Some reports include both eyes in a single image, making whole-image labels unreliable. Using extracted laterality/disease cues and fundus anatomical localization, this module verifies whether the diagnosis applies to one eye. If so, it splits the image into left/right regions, keeps the diagnosed eye, and discards or relabels the other as normal, reducing label noise and improving image--label consistency.

\subsubsection{Reviewer for Quality Control}
To ensure research-grade quality, we add manual QC after automated processing. A stratified random subset is independently reviewed by at least two ophthalmologists for image clarity, anonymization correctness, and image--label consistency. Disagreements are adjudicated by a senior expert, and corrected labels are fed back to improve the pipeline.

Through the collaborative operation of these agents, we achieve a complete processing chain from raw PDF reports to standardized, high-quality ophthalmic classification data. This pipeline not only significantly reduces the workload and error rate of manual curation but also ensures consistency in visual quality, privacy protection, and label accuracy through multi-level validation and quality control.

\begin{figure}[t]
    \centering
    \includegraphics[width=1\linewidth]{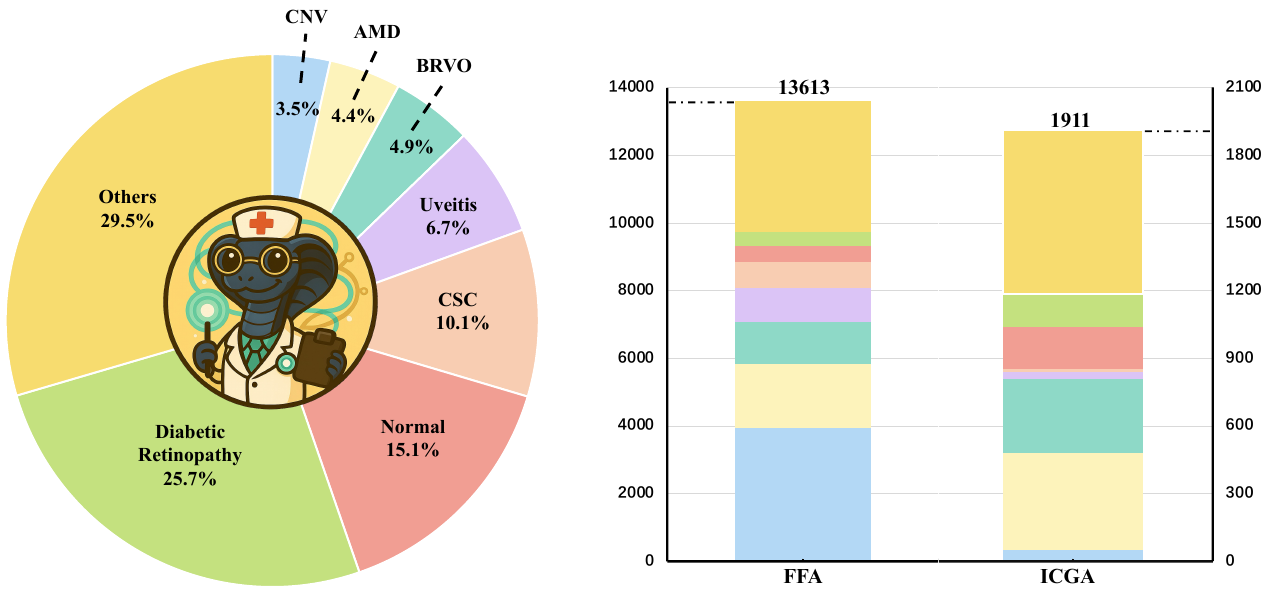}
    \caption{Dataset statistics. (a) Class distribution across 43 ocular categories showing a long-tailed pattern. (b) Modality proportion between FFA and ICGA images.}
    \label{fig:statistics}
    \vspace{-3mm}
\end{figure}

\subsection{Data Stastics}
The final processed dataset consists of 15,524 valid pictures obtained from two angiographic modalities, representing 43 ocular illness categories and one healthy control group.
For model construction, the dataset was divided at the patient level into training and test subsets in an approximate 80\%:20\% ratio.
To guarantee that the disease categories and imaging modalities were distributed similarly across both subsets, a stratified sampling approach was used.

\subsubsection{Data Distribution and Class Imbalance}
The dataset exhibits a pronounced long-tailed class distribution, typical of real-world clinical data. 
Diabetic Retinopathy (DR) is the largest class, with 3,990 photos (25.7\%), followed by Normal (2,338), Central Serous Chorioretinopathy (CSC, 1,572), and Uveitis (1,036).
Together, these high-frequency disorders account for the bulk of samples.
Several categories, however, have fewer than 50 photos, including Punctate Inner Choroidopathy (PIC), Familial Exudative Vitreoretinopathy (FEVR), and Cataract.

This imbalance stems from two factors: (i) true clinical rarity (e.g., PIC and FEVR), which yields few angiographic records, and (ii) imaging-pathway bias, where some common conditions are seldom examined with FFA/ICGA (e.g., cataract, typically diagnosed via slit-lamp). These factors jointly produce a long-tailed distribution. We preserve this real-world skew for realistic evaluation, although it increases training difficulty by biasing models toward majority classes.

\subsubsection{Imaging Modality Analysis}
The dataset includes two main angiographic modalities: fundus fluorescein angiography (FFA) and Indocyanine Green Angiography (ICGA).
A total of 13,613 FFA photos (87.7\%) and 1,911 ICGA images (12.3\%) were obtained, demonstrating that FFA is still the predominate modality in clinical practice.
Most retinal vascular and inflammatory diseases—such as DR, RVO/BRVO/CRVO, and uveitis—are principally scanned with FFA, accounting for more than 95\% of samples, because these disorders are characterised by retinal vascular leakage and perfusion deficits that fluorescein imaging can detect.
In contrast, choroidal and neovascular diseases (e.g., PCV, CNV, and choroidal masses) rely more heavily on ICGA, with proportions typically ranging from 25\% to 50\%, corresponding with clinical usage of ICGA to visualise deep choroidal circulation and neovascular complexes.

\section{Method}
\label{method}
In this section, we describe the design of \ourarch{}. 
We first briefly review the key technical components relevant to angiography classification—efficient spatio-temporal backbones, conditional adaptation, and uncertainty-aware prediction—which form the basis of our framework. 
We then present \ourarch{} and explain how these components are integrated to meet the requirements of ophthalmic angiography classification.

\subsection{Preliminaries}
Before presenting \ourarch{}, we briefly review three building blocks: MedMamba as an efficient SSM-based backbone for capturing local and long-range dependencies, HyperNetworks for lightweight instance-specific adaptation, and Evidential Learning for probabilistic prediction with calibrated uncertainty. Together, they form the basis of our method.

\subsubsection{Medmamba}
MedMamba adopts visual state space models (VSSMs) to model long-range dependencies with linear complexity. 
It uses the 2D selective scan (SS2D) to efficiently propagate information over 2D feature maps, and employs a lightweight gate to fuse scanned features with the original representation.

\begin{figure}[t]
    \centering
    \includegraphics[width=0.95\linewidth]{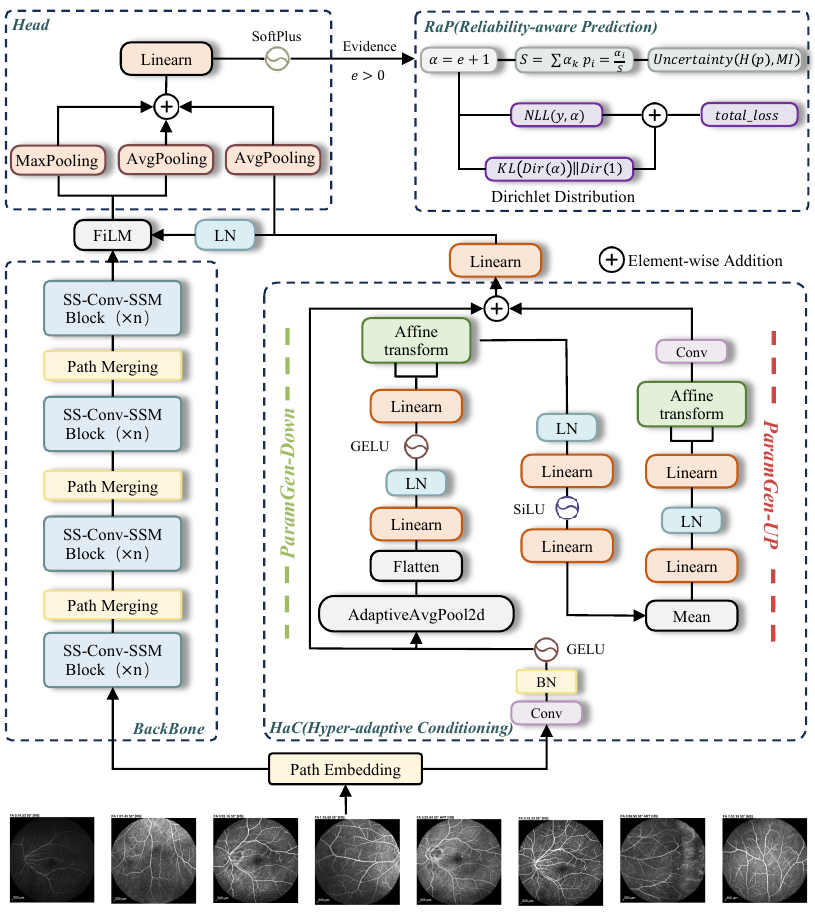}
    \caption{CLEAR-Mamba Framework.}
    \label{fig:ablation}
    \vspace{-3mm}
\end{figure}

Given an input image $x'$, patch embeddings are
\begin{equation}
    X^{(0)}=\mathrm{PatchEmbed}(x')\in\mathbb{R}^{B\times H'\times W'\times D}.
\end{equation}
For each block $\ell=0,\dots,L-1$, SS2D produces global-context features and the gate performs residual fusion:
\begin{equation}
    Y^{(\ell)}=\mathrm{SS2D}(X^{(\ell)}),\quad
    X^{(\ell+1)}=\mathrm{Gate}\!\left([X^{(\ell)},Y^{(\ell)}]\right)+X^{(\ell)} .
\end{equation}
The final representation is obtained by global average pooling:
\begin{equation}
    z=\mathrm{GAP}\!\Big(\mathrm{NHWC\!\to\!NCHW}(X^{(L)})\Big).
\end{equation}

This backbone captures both fine-grained local cues and global structural dependencies, which is crucial for angiography images.

\begin{table*}[t]
    \centering
    \small
    \begin{tabular}{lcccccc}
        \hline
        \textbf{Experiment} & \textbf{Epochs} & \textbf{Batch} & \textbf{Opt.} & \textbf{LR} & \textbf{KL Coef} & \textbf{KL Scale} \\
        \hline
        Primary (Ours) & 150 & 128 & Adam & $10^{-3}$ & $5\times10^{-3}$ & 1.2 \\
        RetinaMNIST & 100 & 96 & Adam & $10^{-3}$ & $5\times10^{-3}$ & 1.2 \\
        OCT-C8 & 20 & 32 & Adam & $10^{-3}$ & $5\times10^{-3}$ & 1.2 \\
        Harvard-GDP & 30 & 32 & Adam & $10^{-3}$ & $5\times10^{-3}$ & 1.2 \\
        \hline
    \end{tabular}
    \caption{Training hyperparameters on our in-house dataset (Primary) and public benchmarks. LR: learning rate; Opt.: optimizer; KL Coef/Scale: coefficient and scaling for the evidential KL term.
    }
    \label{tab:training-setup}
\end{table*}

\subsubsection{HyperNetwork}
A HyperNetwork conditions model parameters on an input $z$:
\begin{equation}
\theta = H_{\psi}(z), \qquad y = F_{\theta}(x),
\end{equation}
thus enabling instance-specific adaptation.

A lightweight form is FiLM, where $G_{\psi}$ predicts feature-wise affine factors:
\begin{equation}
(\gamma,\beta)=G_{\psi}(z), \qquad \tilde{X}=\gamma\odot X+\beta .
\end{equation}
For higher capacity, the hypernetwork can generate (low-rank) adapter parameters:
\begin{equation}
h_{\downarrow}=\phi(W_{\downarrow}h+b_{\downarrow}),\quad
\tilde{h}=h+W_{\uparrow}h_{\downarrow}+b_{\uparrow}.
\end{equation}
We further apply a conditioned gate $\alpha=\sigma(a(z))$ to stabilize the update:
\begin{equation}
X_{\mathrm{out}} = X + \alpha \odot (\tilde{X} - X).
\end{equation}

FiLM is parameter-efficient, while adapter-based hypernetworks provide stronger conditional capacity.

\subsubsection{Evidential Learning}
\label{evidence learning}
Evidential Deep Learning (EDL) models class probabilities with a Dirichlet distribution to provide both prediction and uncertainty. 
Given features $z$, the head outputs non-negative evidence:
\begin{equation}
e=\mathrm{Softplus}(Wz+b),\qquad 
\alpha=e+\mathbf{1},\quad 
\hat{p}_k=\tfrac{\alpha_k}{\sum_{j=1}^{K}\alpha_j}.
\end{equation}
Training combines likelihood with a Dirichlet prior regularizer:
\begin{equation}
\mathcal{L}=\mathcal{L}_{\mathrm{NLL}}(\alpha;y)+
\lambda\,\mathrm{KL}\!\big[\mathrm{Dir}(\alpha)\,\|\,\mathrm{Dir}(\mathbf{1})\big].
\end{equation}
At inference, uncertainty can be summarized by predictive entropy $H(\hat{\mathbf{p}})$ and total evidence $S=\sum_k \alpha_k$ (larger $S$ indicates lower epistemic uncertainty).

\subsection{CLEAR-Mamba}
We propose \ourarch{}, a reliability-enhanced adaptive framework built on MedMamba. It follows an \emph{encode--adapt--predict} pipeline: MedMamba encodes the input into hierarchical features, a lightweight adaptive modulation module applies sample-specific calibration, and an evidential classification head outputs both class probabilities and uncertainty. This design supports high-resolution angiography classification with strong adaptability and reliable uncertainty estimation.

Formally, let $X^{(L)}$ denote the backbone output. 
\ourarch{} applies adaptive modulation:
\begin{equation}
\tilde{X}^{(L)} = X^{(L)} \odot \text{scale} + \text{shift},
\end{equation}
and then feeds $\tilde{X}^{(L)}$ into an evidential classifier that parameterizes a Dirichlet distribution:
\begin{equation}
\alpha = e+\mathbf{1},\qquad S=\sum_{k=1}^K \alpha_k,\qquad \hat{p}_k=\frac{\alpha_k}{S}.
\end{equation}

\subsubsection{Hyper-adaptive Conditioning} 
A HyperNetwork conditions model parameters on an input $z$:
\begin{equation}
\theta = H_{\psi}(z), \qquad y = F_{\theta}(x),
\end{equation}
thus enabling instance-specific adaptation.

A lightweight form is FiLM, where $G_{\psi}$ predicts feature-wise affine factors:
\begin{equation}
(\gamma,\beta)=G_{\psi}(z), \qquad \tilde{X}=\gamma\odot X+\beta .
\end{equation}
For higher capacity, the hypernetwork can generate (low-rank) adapter parameters:
\begin{equation}
h_{\downarrow}=\phi(W_{\downarrow}h+b_{\downarrow}),\quad
\tilde{h}=h+W_{\uparrow}h_{\downarrow}+b_{\uparrow}.
\end{equation}
We further apply a conditioned gate $\alpha=\sigma(a(z))$ to stabilize the update:
\begin{equation}
X_{\mathrm{out}} = X + \alpha \odot (\tilde{X} - X).
\end{equation}

FiLM is parameter-efficient, while adapter-based hypernetworks provide stronger conditional capacity.

\subsubsection{Reliability-aware Prediction} 
To endow \ourarch{} with reliable decision making, we integrate the evidential formulation (see Sec.~\ref{evidence learning}) into the final classifier. 
Instead of outputting deterministic logits, the \textbf{RaP} head produces Dirichlet parameters $\alpha$ from the backbone features, yielding both class probabilities and an associated measure of uncertainty. 

During training, we adopt the evidential objective introduced in Preliminaries, i.e., the marginal likelihood with a KL regularizer toward the non-informative prior $\mathrm{Dir}(\mathbf{1})$. 
This encourages the model to express uncertainty when evidence is insufficient. 

At inference, the \textbf{RaP} head outputs calibrated probabilities $\hat{\mathbf p}$ as well as uncertainty summaries. 
A key metric is the predictive entropy
\begin{equation}
H(\hat{\mathbf p}) = -\sum_{k=1}^{K} \hat{p}_k \log \hat{p}_k,
\end{equation}
which grows when the prediction is uncertain. 
These scores provide a practical basis for selective review and risk-aware deployment in clinical settings.
\section{Experiment}
\label{exp}

\begin{table*}[t]
    \centering
    \begin{tabular}{c|c|c|c|c|c|c|c|c}
        \toprule
        \textbf{Type} & \textbf{Model} & {\textbf{Para. (M)}} & {\textbf{P(\%)↑}} & {\textbf{Se(\%)↑}} & {\textbf{Sp(\%)↑}} & {\textbf{F1(\%)↑}} & {\textbf{OA(\%)↑}} & {\textbf{AUC↑}} \\
        \midrule
        \multirow{8}{*}[-0.8ex]{\centering\textbf{General Model}}
        & ResNet18(28) & 34 & 7.77 & 7.18 & 98.31 & 6.52 & 39.08 & 0.7389 \\
        & ResNet18(224) & 34 & 18.01 & 14.41 & 98.71 & 14.01 & 51.67 & 0.8633 \\
        & ResNet50(28) & 71 & 4.41 & 5.22 & 98.15 & 4.16 & 35.65 & 0.6795 \\
        & ResNet50(224) & 71 & 18.94 & 15.05 & 98.74 & 15.55 & 51.47 & 0.8596 \\
        & DINOv3-ViT-B & 86 & 14.46 & 12.03 & 98.59 & 12.24 & 47.66 & 0.8176 \\
        & DINOv3-ViT-H+ & 840 & 10.91 & 8.87 & 98.44 & 8.72 & 42.66 & 0.7702 \\
        & DINOv3-ConvNeXt-B & 89 & 14.79 & 11.10 & 98.56 & 11.24 & 46.83 & 0.8173 \\
        & DINOv3-ConvNeXt-L & 198 & 15.69 & 11.25 & 98.61 & 10.96 & 46.93 & 0.8272 \\
        & MambaVision-T	& 150	&7.56	&6.37	&98.33	&5.34	&41.48	&0.7620 \\
        & MambaVision-S	& 95	&10.33	&8.65	&98.53	&7.74	&46.57	&0.8504 \\
        & MambaVision-B	& 292	&9.57	&9.17	&98.53	&8.26	&46.76	&0.8143 \\
        & MambaVision-L	& 684	&10.94	&8.45	&98.49	&7.67	&45.8	&0.8136 \\
        \midrule
        \multirow{7}{*}[-0.6ex]{\centering\textbf{Medical Model}}
        & MedViT\_V2\_tiny & 32 & 18.44 & 15.91 & 98.84 & 15.25 & 55.29 & 0.8784 \\
        & MedViT\_V2\_small & 83 & 18.79 & 14.55 & 98.77 & 14.26 & 53.11 & 0.8632 \\
        & MedViT\_V2\_base & 187 & 18.01 & 13.46 & 98.72 & 13.31 & 51.92 & 0.8595 \\
        & MedViT\_V2\_large & 330 & 18.80 & 16.41 & 98.78 & 16.83 & 53.04 & 0.8647 \\
        & Medmamba-T	& 15 & 19.22 & 17.08 & 98.79	& 16.87	& 53.71	& 0.8714 \\
        & Medmamba-S	& 19 & 17.91 & 14.83 & 98.74	& 14.43	& 51.92	& 0.8614 \\
        & Medmamba-B	& 40 & 22.58 & 18.68 & 98.85	& 18.34	& 55.38	& 0.8790 \\
        & Medmamba-X	& 15 & 20.12 & 15.61 & 98.82 & 14.97 & 54.83 & 0.8833 \\
        & \textbf{\textit{CLEAR-T}}	& 15	& 24.05	& 20.59	& 98.94	& 20.45	& 59.26	& 0.8450 \\
        & \textbf{\textit{CLEAR-S}}	& 19	& 24.52	& 21.54	& 98.94	& 20.85	& 58.97	& 0.8590 \\
        & \textbf{\textit{CLEAR-B}}	& 40	& 27.66	& 22.45	& 98.95	& 22.71 & 59.06 & 0.8360 \\
        \bottomrule
    \end{tabular}
    \caption{Quantitative comparison of general-purpose, medical-specific, and proposed models on the in-house angiography dataset.All models are trained and evaluated under identical experimental settings for fair comparison. The proposed \ourarch{} framework is presented in three scales (\textit{T/S/B}); for convenience, it is referred to as \textbf{\textit{CLEAR}} in the following analysis.}
    \label{tab:performance-comparison}
\end{table*}

This section evaluates \ourarch{} for ophthalmic image classification. We first describe the setup (datasets, metrics, and tuned hyper-parameters), then compare with baselines and assess robustness across datasets, followed by results and analysis.

\subsection{Implementation details}
All experiments were implemented in PyTorch 2.8 and run on a single NVIDIA GeForce RTX 5090 GPU. 
For public datasets, we followed the official preprocessing/protocols, and trained baselines with their default settings. 
\ourarch{} used the hyperparameters in Table~\ref{tab:training-setup}; otherwise, we keep a unified default configuration.

Specifically, the \textbf{RaP} (Reliability-aware Prediction) module was operated in the \textit{adaptive} mode, 
with a KL divergence coefficient of $5\times10^{-3}$ and a scaling factor of $1.2$. 
For the \textbf{HaC} (Hyper-adaptive Conditioning) module, we systematically explored two key parameters: 
the reduction ratio $\{1, 2, 4, 8\}$ and the hidden feature dimension $\{32, 64, 96, 128\}$, 
resulting in 16 experimental configurations to comprehensively assess their influence on performance.

\subsection{Experimental Datasets}
\label{subsec:experimental datasets}
This study targets multi-disease classification on FFA/ICGA sequences, where public datasets are scarce. Most existing ophthalmic benchmarks focus on single-disease tasks (e.g., DR or glaucoma) due to data sensitivity and privacy constraints. We therefore construct an in-house multi-disease angiography dataset with temporal information covering 43 categories as our primary benchmark. To further evaluate robustness and generalization, we also test on several external single-disease datasets with related clinical settings, described below.

\subsubsection{RetinaMNIST}
RetinaMNIST is a MedMNIST subset~\cite{yang2021medmnist} derived from DeepDRiD~\cite{liu2022deepdrid}, containing 1,600 fundus images labeled into five diabetic retinopathy severity levels. Images are center-cropped and resized to $28\times28$, with splits of 1,080/120/400 for train/val/test. The task is ordinal DR grading, and remains challenging due to fine-grained differences and imbalanced severity distribution.

\subsubsection{Harvard-GDP}
Harvard-GDP is a subset of the Harvard Glaucoma Detection and Progression dataset~\cite{luo2023harvard}, collected from 1,000 patients. Each sample is a $224\times224$ RNFL thickness (RNFLT) map derived from 3D OCT, labeled as glaucomatous vs. non-glaucomatous based on visual-field–verified diagnosis. Using the provided train/val/test protocol, this benchmark is challenging because class differences in RNFLT maps are often subtle.

\subsubsection{OCT-C8}
OCT-C8~\cite{subramanian2022classification} is a composite OCT B-scan dataset compiled from public sources (e.g., Kaggle and Open-ICPSR), with $\sim$24k images across eight classes (AMD, CNV, DRUSEN, DME, DR, MH, CSR, Normal). Images are resized to $224\times224$ and split 75\%/15\%/15\% for train/val/test. It serves as a multi-class retinal OCT benchmark with high intra-class variability and subtle inter-class differences.

\begin{figure*}[t]
    \centering
    \includegraphics[width=1\linewidth]{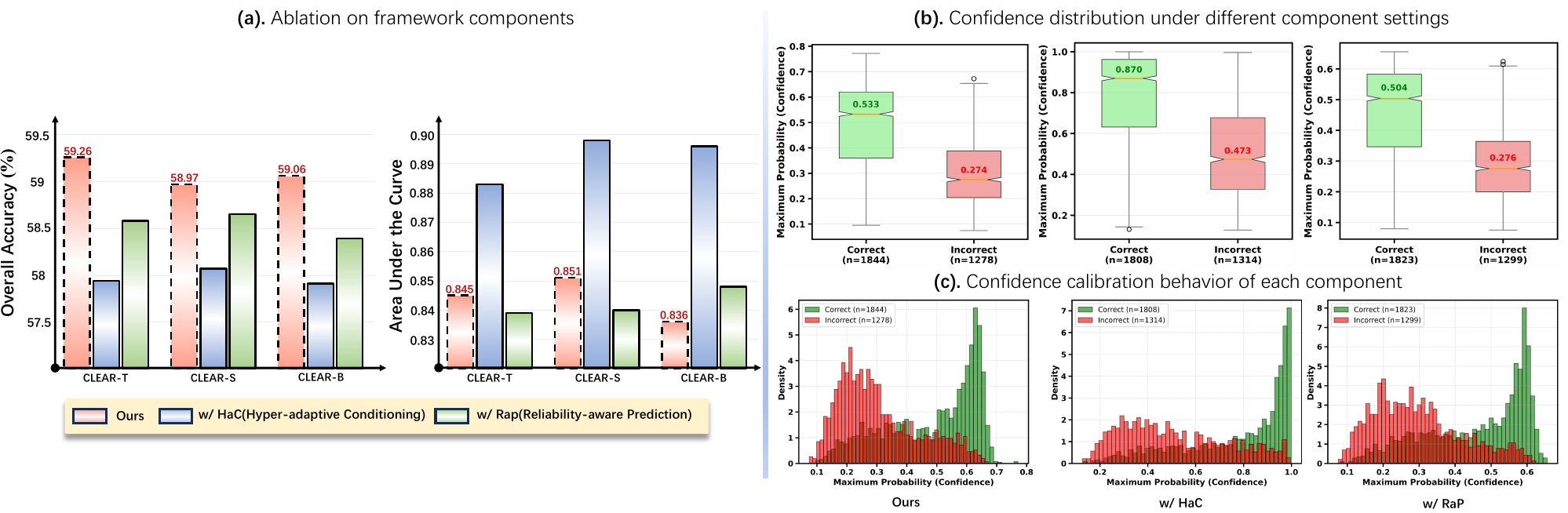}
    \caption{Component-level ablation study of the \model{CLEAR} framework.
(a) Comparison of overall accuracy (OA) and AUC among different component configurations.
(b) Boxplots and (c) density distributions of prediction confidence for correctly and incorrectly classified samples in different variants (\textbf{HaC}, \textbf{RaP}, and full model).}
    \label{fig:ablation_1}
    \vspace{-3mm}
\end{figure*}

\subsection{Evaluation Metrics}
\label{sec:metrics}
We consider a $K$-class classification problem on $\mathcal{D}=\{(x_i,y_i)\}_{i=1}^N$, where $y_i\in\{1,\dots,K\}$. 
The model outputs $\mathbf{p}_i\in[0,1]^K$ and $\hat{y}_i=\arg\max_k p_{i,k}$. 
Let $\mathrm{TP}_k,\mathrm{FP}_k,\mathrm{FN}_k,\mathrm{TN}_k$ denote the confusion counts for class $k$.

We report Overall Accuracy (OA) and macro-averaged Precision, Sensitivity (Recall), Specificity, F1-score, and AUC. 
Specifically,
\begin{align}
\mathrm{OA} &= \frac{\sum_{k=1}^{K}\mathrm{TP}_k}{N}, \label{eq:oa}\\
\mathrm{Precision}_k &= \frac{\mathrm{TP}_k}{\mathrm{TP}_k+\mathrm{FP}_k}, \label{eq:prec}\\
\mathrm{Sensitivity}_k &= \frac{\mathrm{TP}_k}{\mathrm{TP}_k+\mathrm{FN}_k}, \label{eq:recall}\\
\mathrm{Specificity}_k &= \frac{\mathrm{TN}_k}{\mathrm{TN}_k+\mathrm{FP}_k}, \label{eq:spec}\\
\mathrm{F1}_k &= \frac{2\,\mathrm{Precision}_k\,\mathrm{Sensitivity}_k}
{\mathrm{Precision}_k+\mathrm{Sensitivity}_k}. \label{eq:f1}
\end{align}
Macro averages are computed as $\frac{1}{K}\sum_{k=1}^{K}(\cdot)$. 
AUC is computed in a one-vs-rest manner for each class and then macro-averaged.

In addition, we also analyzed the model’s computational complexity in terms of floating-point operations (FLOPs) and the total number of learnable parameters, 
providing a complementary view of efficiency alongside predictive performance.

\subsection{Results on the In-house FFA/ICGA Dataset}
We evaluate \ourarch{} on our in-house multi-disease angiography dataset with full FFA/ICGA temporal sequences. This experiment tests its ability to model temporal hemodynamic cues and handle heterogeneous lesions. We further compare against representative CNN-, Transformer-, and Mamba-based baselines to assess reliability calibration and generalization under real-world clinical variability.

\subsubsection{The Data}
We primarily evaluate \ourarch{} on the test split of our in-house multi-disease angiography dataset, which serves as the main benchmark in this study. Detailed are provided in Sec.~\ref{dataset}.

\subsubsection{Preprocessing and Partitioning}
\label{sec:data_preprocess}
All FFA/ICGA frames were resized to $224\times224$. During training, we applied random resized cropping and horizontal flipping, followed by normalization (mean/std $=(0.5,0.5,0.5)$). For validation, we used only resizing and normalization. We split the in-house dataset into 80\%/20\% for train/test. The best checkpoint was selected by validation OA and evaluated on the test set. For public datasets, we follow their official preprocessing and protocols for fair comparison.

\subsubsection{Overall Comparison with Existing Models}
\label{sec:overall-comparison}
To comprehensively evaluate the discriminative capability and domain adaptability of our framework, 
we benchmarked the proposed \textbf{\textit{CLEAR-Mamba}} models at three scales (\textit{T/S/B}) under identical experimental settings. 
For simplicity, we use the term \textbf{\textit{CLEAR}} to refer collectively to these variants in the following discussion.

As shown in Table~\ref{tab:performance-comparison}, general-purpose backbones (e.g., ResNet, DINOv3, MambaVision) degrade substantially when trained from scratch on our angiography data. Their strong performance on natural images largely relies on large-scale pretraining; with random initialization and limited medical supervision, they struggle to capture fine-grained vascular textures and lesion cues. For instance, ResNet18(224) achieves 51.67\% OA / 14.01\% F1, and DINOv3-ViT-B obtains 47.66\% OA / 12.24\% F1.
In contrast, medical-oriented models such as MedViT-V2 and MedMamba are more robust when trained from scratch. Their domain-aware designs (e.g., hierarchical fusion and context-preserving state-space modeling) provide stronger inductive bias, enabling effective learning on moderate-scale medical data. As a result, they achieve higher OA (51--55\%) and F1 (13--18\%) without large-scale pretraining.

Compared with MedViT-V2 and MedMamba, \textbf{\textit{CLEAR}} achieves consistently higher sensitivity and OA across all model scales. 
\model{CLEAR-T/S/B} obtain F1 scores of 20.45\%, 20.85\%, and 22.71\%, with OA of 59.26\%, 58.97\%, and 59.06\%, respectively, while maintaining high specificity (98.94--98.95\%). 
Overall, CLEAR improves F1 by about +6--8 points and OA by about +4 points over MedMamba-X. 
We attribute these gains to \textbf{HaC} for instance-conditioned feature modulation and \textbf{RaP} for evidential uncertainty modeling, yielding more stable and calibrated predictions.

\begin{figure}[t]
\centering
\includegraphics[width=1\linewidth]{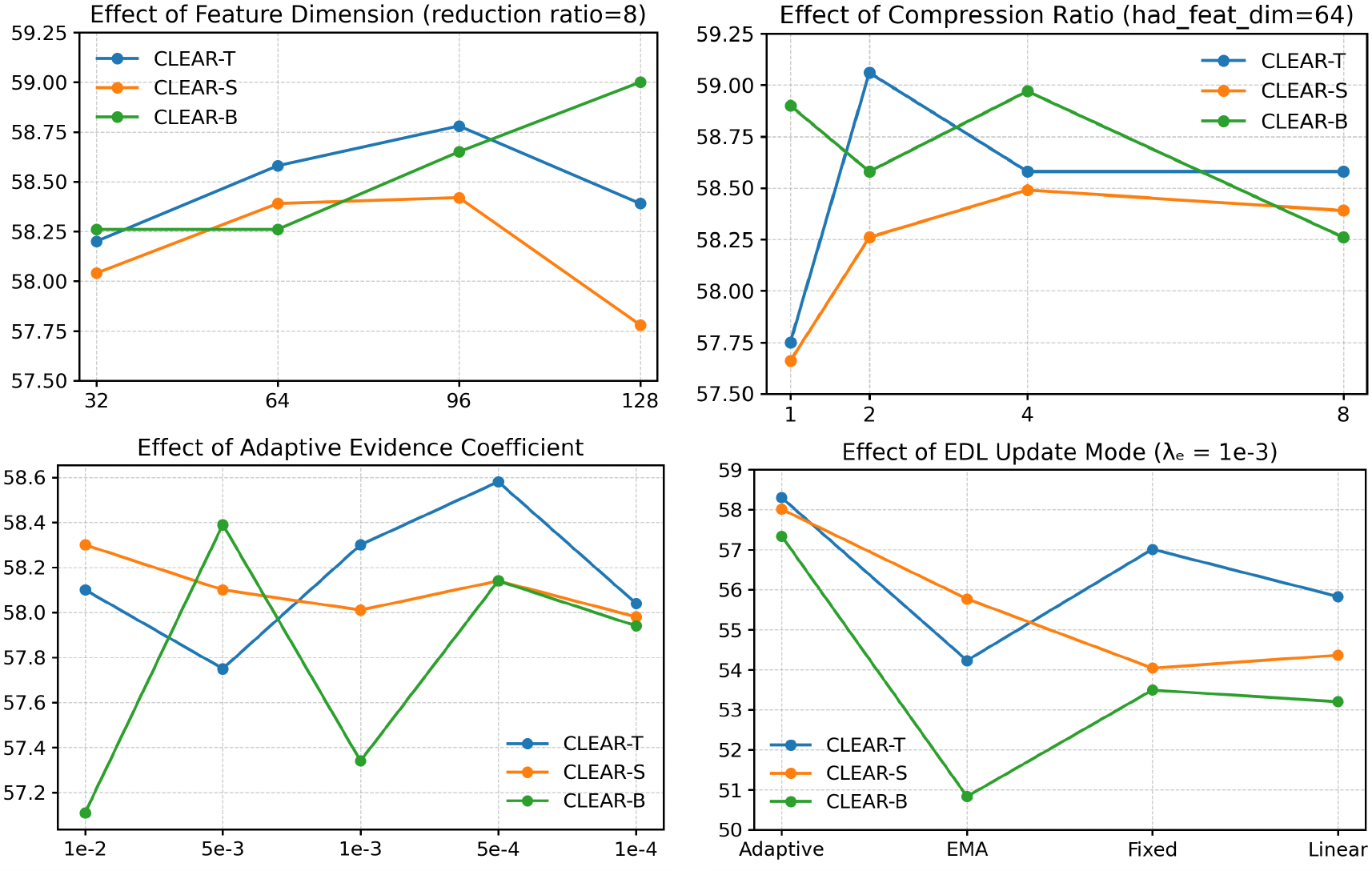}
\caption{Hyperparameter-level ablation study of the \model{CLEAR} framework. Subfigures (a–d) are arranged clockwise from the top-left. 
(a) Effect of the hypernetwork feature dimension (\textit{had\_feat\_dim}); 
(b) impact of the compression ratio (\textit{r}); 
(c) sensitivity to the adaptive evidence coefficient ($\lambda_e$); and 
(d) comparison of EDL update strategies.}
\label{fig:ablation_2}
\end{figure}

\subsubsection{Ablation Study and Comparisons}
We conduct ablation studies to assess each component and hyperparameter effects. As shown in Fig.~\ref{fig:ablation_1}, the full model (“Ours”) achieves the best OA and AUC across scales. Both \textbf{HaC} and \textbf{RaP} improve performance, and combining them yields complementary gains.

\textbf{Components Ablation.}
For clarity, we report \textbf{\textit{CLEAR-B}} as a representative case (Fig.~\ref{fig:ablation_1}(b)--(c)), with similar trends on \textbf{\textit{CLEAR-T/S}}. 
We observe three behaviors from the confidence distributions. 
\textbf{+HaC} is \textit{overconfident}, assigning high confidence to both correct and incorrect predictions (median 0.870 vs. 0.473), which harms OA. 
\textbf{+RaP} is \textit{overly conservative}, with confidence concentrated around $\sim$0.6 even for easy cases. 
The full model achieves better calibration and separation, with median confidences of 0.533 (correct) and 0.274 (incorrect), yielding a larger confidence gap than +RaP and explaining its superior OA/AUC.

\textbf{Impact of Hyperparameters.}
We analyze the sensitivity of \textbf{HaC} and \textbf{RaP} to key hyperparameters (Fig.~\ref{fig:ablation_2}), including \texttt{had\_feat\_dim}, the reduction ratio $r$, the evidential coefficient $\lambda_e$, and the EDL update strategy. 
For \textbf{HaC}, increasing \texttt{had\_feat\_dim} improves OA but quickly saturates (and may slightly drop) beyond 128, indicating diminishing returns and potential over-conditioning, especially for smaller variants (\model{CLEAR-T/S}); \model{CLEAR-B} benefits more but becomes slightly sensitive under stronger compression (higher $r$). 
For \textbf{RaP}, OA is stable across $\lambda_e\in[10^{-4},10^{-2}]$ (within $\sim$1\%, Fig.~\ref{fig:ablation_2}(c)), while the \textit{Adaptive} update consistently yields the best results across scales (Fig.~\ref{fig:ablation_2}(d)), highlighting the value of dynamic evidence weighting for reliable uncertainty modeling.

\subsubsection{In-depth Analysis}
Beyond feature separability, we further examine the reliability of model predictions by analyzing confidence and uncertainty under different ambiguity levels.

\begin{figure}[t]
    \centering
    \includegraphics[width=0.95\linewidth]{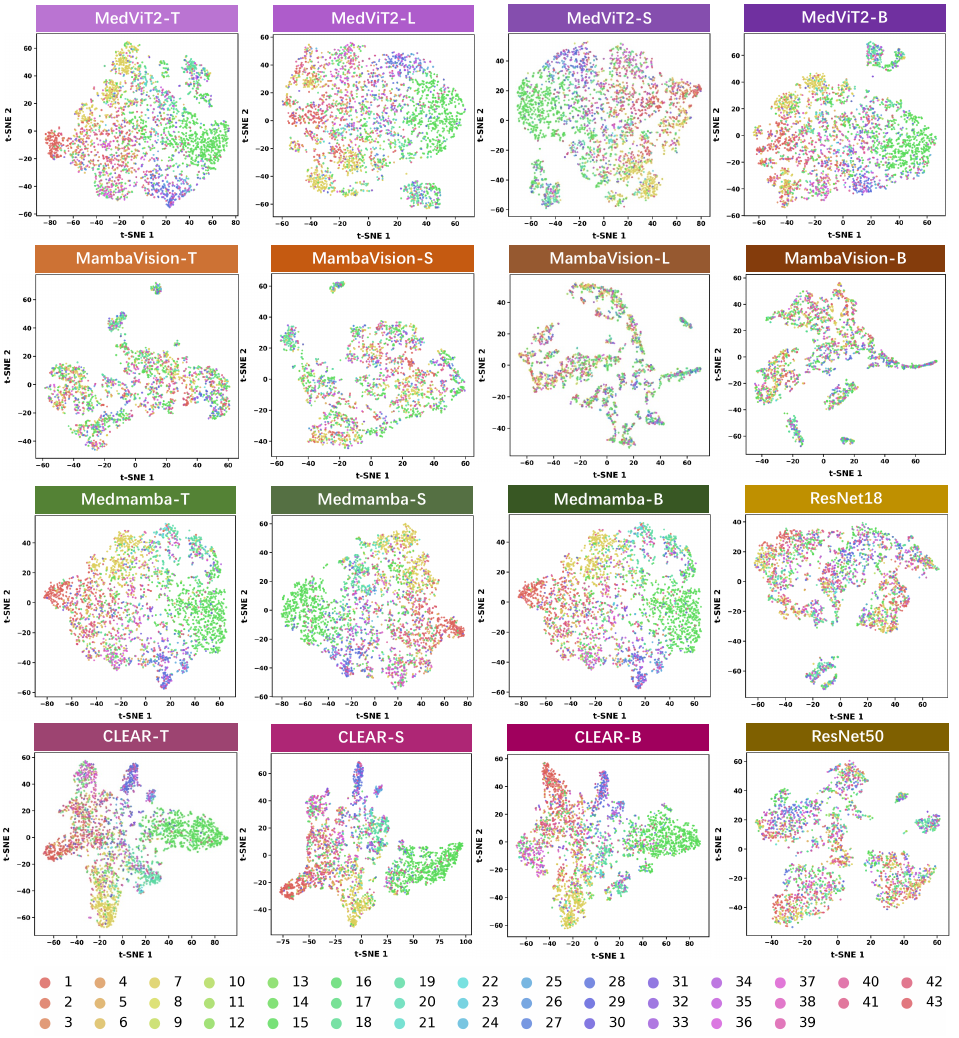}
    \caption{t-SNE visualization of feature embeddings from ResNet, MedViT2, MambaVision, MedMamba, and \model{CLEAR} on the in-house dataset.}
    \label{fig:t-sne}
    \vspace{-3mm}
\end{figure}

\textbf{T-SNE Feature Visualization.}
We further visualize learned embeddings with t-SNE (Fig.~\ref{fig:t-sne}). \model{CLEAR} shows tighter intra-class clustering and clearer inter-class separation than ResNet, MedViT, MambaVision, and MedMamba, even under the challenging 43-class setting. In contrast, other models exhibit more overlap, indicating weaker feature discriminability.

\textbf{Confidence-Aware Reliability Analysis.}
We illustrate three samples with progressively increasing uncertainty. A confident and reliable prediction is observed for \textit{BRVO}, characterized by a peaked distribution (top-1 = 0.63) and low uncertainty (Total Unc $\approx$ 1.84). For a \textit{DR} sample, the prediction remains correct but becomes less decisive (top-1 = 0.40; others $\approx$ 0.10) with higher uncertainty ($\approx$ 2.79), indicating that manual review may be needed. In a more ambiguous \textit{uveitis} example, the model misclassifies it as \textit{retinal hemorrhage} with low confidence (top-1 = 0.17) and high uncertainty ($\approx$ 3.49), suggesting that \model{CLEAR} is aware of its unreliability rather than making an over-confident error.

\begin{figure}[t]
    \centering
    \includegraphics[width=0.95\linewidth]{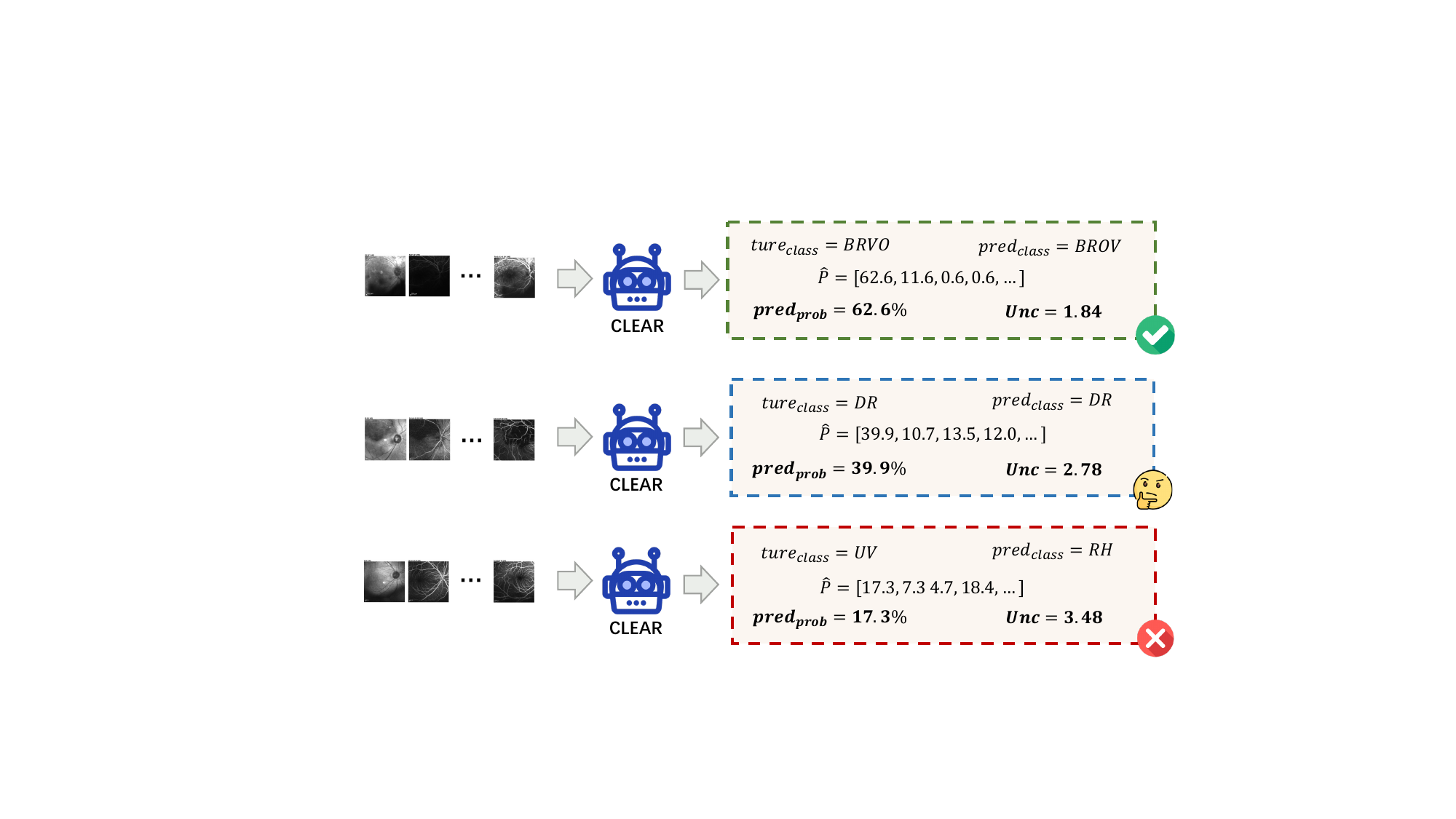}
    \caption{Case studies of the \textbf{RaP} module in \model{CLEAR}. Representative examples showing how quantifies uncertainty and adjusts confidence under different prediction scenarios: (a) confident and correct, (b) cautious but correct, and (c) prudent failure with high uncertainty. }
    \label{fig:case}
    \vspace{-3mm}
\end{figure}

\subsection{Results on Public Datasets}
Beyond our in-house angiography dataset, we evaluate cross-domain robustness on three public retinal benchmarks—RetinaMNIST, OCT-C8, and Harvard-GDP—covering fundus, OCT, and volumetric glaucoma data. These datasets allow us to assess whether \ourarch{} maintains stable performance under domain shift and limited data, and to verify its scalability across input resolutions and label granularities without modifying the architecture or hyperparameters.

\subsubsection{The Data}
The details of RetinaMNIST, Harvard-GDP, and OCT-C8 datasets are provided in Section~\ref{subsec:experimental datasets}.

\begin{table}[t]
\centering
\caption{Performance comparison on the \textbf{Harvard-GDP} for the glaucoma \textbf{TD Progression forecasting} task. Values before and after the slash ("/") indicate the results under \textbf{single-modality} and \textbf{multimodal fusion} settings, respectively.}
\begin{tabular*}{0.85\linewidth}{@{\extracolsep{\fill}} lcc @{}}
\toprule
\textbf{Model} & \textbf{Acc$\uparrow$} & \textbf{AUC$\uparrow$} \\
\midrule
\midrule
VGG & 0.78/0.80 & 0.84/0.79 \\
ResNet & 0.75/0.74 & 0.74/0.75 \\
ResNext & 0.78/0.77 & 0.78/0.76 \\
WideResNet & 0.77/0.79 & 0.77/0.80 \\
EfficientNet & 0.73/0.78 & 0.76/0.79 \\
ConvNext & 0.74/0.81 & 0.78/0.81 \\
ViT & 0.73/0.77 & 0.68/0.79 \\
Swin & 0.71/0.77 & 0.68/0.77 \\
\midrule
\model{CLEAR} & \textbf{0.91} & \textbf{0.85} \\
\bottomrule
\bottomrule
\end{tabular*}
\label{tab:progression_forecasting}
\end{table}

\begin{table}[t]
\centering
\caption{Performance comparison on the \textbf{OCT-C8} dataset, which includes eight categories of retinal diseases (CNV, DME, DRUSEN, NORMAL, CSC, MH, AMD, and RVO). }
\setlength{\tabcolsep}{8pt}
\begin{tabular*}{0.85\linewidth}{@{\extracolsep{\fill}} lcc @{}}
\toprule
\textbf{Model} & \textbf{OA$\uparrow$} & \textbf{AUC$\uparrow$} \\
\midrule
\midrule
DenseNet201    & 0.9214 & -- \\
VGG19          & 0.9018 & -- \\
InceptionV3    & 0.8675 & -- \\
\model{CLEAR-T} & 0.9254 & 0.9918 \\
\model{CLEAR-S} & \textbf{0.9450} & \textbf{0.9961} \\
\model{CLEAR-B} & 0.9321 & 0.9934 \\
\bottomrule
\bottomrule
\end{tabular*}
\label{tab:octc8_results}
\end{table}

\subsubsection{Results on Public Retinal Datasets}

\begin{table}[t]
\centering
\caption{Performance comparison on the \textbf{RetinaMNIST}. Values indicate the overall accuracy (OA) and AUC on the test set.}
\setlength{\tabcolsep}{8pt}       
\begin{tabular*}{0.9\linewidth}{@{\extracolsep{\fill}} lcc @{}}
\toprule
\textbf{Model} & \textbf{OA$\uparrow$} & \textbf{AUC$\uparrow$} \\
\midrule
\midrule
ResNet18 (28)   & 52.4  & 0.717 \\
ResNet18 (224)  & 49.3  & 0.710 \\
ResNet50 (28)   & 52.8  & 0.726 \\
ResNet50 (224)  & 51.1  & 0.716 \\
Auto-sklearn    & 51.5  & 0.690 \\
AutoKeras       & 50.3  & 0.719 \\
Google AutoML   & 53.1  & 0.750 \\
MedViT-T        & 53.4  & 0.752 \\
MedViT-S        & 56.1  & 0.773 \\
MedViT-L        & 55.2  & 0.754 \\
MedMamba-T      & 54.3  & 0.747 \\
MedMamba-S      & 54.5  & 0.718 \\
MedMamba-B      & 55.3  & 0.715 \\
\model{CLEAR-T} & \textbf{55.8} & \textbf{0.738} \\
\model{CLEAR-S} & \textbf{56.5} & \textbf{0.729} \\
\model{CLEAR-B} & \textbf{56.8} & \textbf{0.742} \\
\bottomrule
\bottomrule
\end{tabular*}
\label{tab:retinamnist_results}
\end{table}

\textbf{Results on the Harvard-GDP.} 
Table~\ref{tab:progression_forecasting} reports results on Harvard-GDP. Due to setting differences, we only compare the progression forecasting results with those reported in~\cite{luo2023harvard}. As class imbalance reduces the sensitivity of metrics to model differences, we present results on the TD Progression task only and refer readers to~\cite{luo2023harvard} for full settings and additional results. Notably, \model{CLEAR} achieves state-of-the-art performance even with a single modality, outperforming all reported multimodal fusion models, suggesting strong progression cue modeling from OCT alone.

\textbf{Results on the OCT-C8.} 
Table~\ref{tab:octc8_results} summarizes multi-class classification results on OCT-C8. Under the same protocol, \model{CLEAR} consistently improves OA and AUC. In particular, \model{CLEAR-S} achieves the best overall performance (94.5\% OA, 0.9961 AUC), surpassing both CNN-based and hybrid medical vision baselines.

\textbf{Results on the RetinaMNIST.}
Table~\ref{tab:retinamnist_results} presents results on RetinaMNIST, where baselines are reproduced from the MedMamba paper. Under the same setting, \model{CLEAR} variants consistently outperform all baselines; \model{CLEAR-B} achieves the best accuracy (56.8\%) and AUC (0.742), demonstrating robust generalization on retinal disease classification.
\section{Conclusion}
We proposed \ourarch{}, a reliability-enhanced and adaptive framework for single-modality, multi-sequence ophthalmic angiography classification. It combines (1) a MedMamba backbone for efficient spatio-temporal modeling, (2) a Hyper-adaptive Conditioning (HaC) module for lightweight sample-wise adaptation, and (3) a Reliability-aware Prediction (RaP) head based on evidential learning to produce calibrated predictions with uncertainty estimates. 
We curate a large-scale angiography dataset covering 43 diseases with complete FFA/ICGA sequences, and evaluate our method both in-house and on three public benchmarks (RetinaMNIST, OCT-C8, and Harvard-GDP). 
CLEAR-Mamba consistently outperforms CNN/ViT/Mamba baselines in OA/F1/AUC, while exhibiting improved calibration and robustness under heterogeneous and imbalanced conditions, supporting its potential for trustworthy clinical use. 
Future work will focus on multi-center prospective validation, extending to multimodal fusion, integrating uncertainty with selective prediction and human-in-the-loop review, and exploring sequence-aware pretraining to further improve real-world generalization.

\bibliography{main}

@article{dao2025recent,
  title={Recent advances in medical image classification},
  author={Dao, Loan and Ly, Ngoc Quoc},
  journal={arXiv preprint arXiv:2506.04129},
  year={2025}
}

@article{dosovitskiy2020image,
  title={An image is worth 16x16 words: Transformers for image recognition at scale},
  author={Dosovitskiy, Alexey and Beyer, Lucas and Kolesnikov, Alexander and Weissenborn, Dirk and Zhai, Xiaohua and Unterthiner, Thomas and Dehghani, Mostafa and Minderer, Matthias and Heigold, Georg and Gelly, Sylvain and others},
  journal={arXiv preprint arXiv:2010.11929},
  year={2020}
}

@article{manzari2023medvit,
  title={MedViT: a robust vision transformer for generalized medical image classification},
  author={Manzari, Omid Nejati and Ahmadabadi, Hamid and Kashiani, Hossein and Shokouhi, Shahriar B and Ayatollahi, Ahmad},
  journal={Computers in biology and medicine},
  volume={157},
  pages={106791},
  year={2023},
  publisher={Elsevier}
}

@article{lecun2002gradient,
  title={Gradient-based learning applied to document recognition},
  author={LeCun, Yann and Bottou, L{\'e}on and Bengio, Yoshua and Haffner, Patrick},
  journal={Proceedings of the IEEE},
  volume={86},
  number={11},
  pages={2278--2324},
  year={2002},
  publisher={Ieee}
}

@article{yue2024medmamba,
  title={Medmamba: Vision mamba for medical image classification},
  author={Yue, Yubiao and Li, Zhenzhang},
  journal={arXiv preprint arXiv:2403.03849},
  year={2024}
}

@book{gu2023modeling,
  title={Modeling sequences with structured state spaces},
  author={Gu, Albert},
  year={2023},
  publisher={Stanford University}
}

@article{luo2025survey,
  title={A Survey of Multimodal Ophthalmic Diagnostics: From Task-Specific Approaches to Foundational Models},
  author={Luo, Xiaoling and Zheng, Ruli and Zheng, Qiaojian and Du, Zibo and Yang, Shuo and Ding, Meidan and Xu, Qihao and Liu, Chengliang and Shen, Linlin},
  journal={arXiv preprint arXiv:2508.03734},
  year={2025}
}

@article{chen2021multimodal,
  title={Multimodal, multitask, multiattention (M3) deep learning detection of reticular pseudodrusen: Toward automated and accessible classification of age-related macular degeneration},
  author={Chen, Qingyu and Keenan, Tiarnan DL and Allot, Alexis and Peng, Yifan and Agr{\'o}n, Elvira and Domalpally, Amitha and Klaver, Caroline CW and Luttikhuizen, Daniel T and Colyer, Marcus H and Cukras, Catherine A and others},
  journal={Journal of the American Medical Informatics Association},
  volume={28},
  number={6},
  pages={1135--1148},
  year={2021},
  publisher={Oxford Academic}
}

@article{mehta2021automated,
  title={Automated detection of glaucoma with interpretable machine learning using clinical data and multimodal retinal images},
  author={Mehta, Parmita and Petersen, Christine A and Wen, Joanne C and Banitt, Michael R and Chen, Philip P and Bojikian, Karine D and Egan, Catherine and Lee, Su-In and Balazinska, Magdalena and Lee, Aaron Y and others},
  journal={American Journal of Ophthalmology},
  volume={231},
  pages={154--169},
  year={2021},
  publisher={Elsevier}
}

@article{hua2020convolutional,
  title={Convolutional network with twofold feature augmentation for diabetic retinopathy recognition from multi-modal images},
  author={Hua, Cam-Hao and Kim, Kiyoung and Huynh-The, Thien and You, Jong In and Yu, Seung-Young and Le-Tien, Thuong and Bae, Sung-Ho and Lee, Sungyoung},
  journal={IEEE Journal of Biomedical and Health Informatics},
  volume={25},
  number={7},
  pages={2686--2697},
  year={2020},
  publisher={IEEE}
}

@article{he2021multi,
  title={Multi-modal retinal image classification with modality-specific attention network},
  author={He, Xingxin and Deng, Ying and Fang, Leyuan and Peng, Qinghua},
  journal={IEEE transactions on medical imaging},
  volume={40},
  number={6},
  pages={1591--1602},
  year={2021},
  publisher={IEEE}
}

@article{zou2024confidence,
  title={Confidence-aware multi-modality learning for eye disease screening},
  author={Zou, Ke and Lin, Tian and Han, Zongbo and Wang, Meng and Yuan, Xuedong and Chen, Haoyu and Zhang, Changqing and Shen, Xiaojing and Fu, Huazhu},
  journal={Medical Image Analysis},
  volume={96},
  pages={103214},
  year={2024},
  publisher={Elsevier}
}

@article{study1999age,
  title={The age-related eye disease study (AREDS): design implications AREDS report no. 1},
  author={Study, The Age-Related Eye Disease and others},
  journal={Controlled clinical trials},
  volume={20},
  number={6},
  pages={573--600},
  year={1999},
  publisher={Elsevier}
}

@article{liu2022deepdrid,
  title={Deepdrid: Diabetic retinopathy—grading and image quality estimation challenge},
  author={Liu, Ruhan and Wang, Xiangning and Wu, Qiang and Dai, Ling and Fang, Xi and Yan, Tao and Son, Jaemin and Tang, Shiqi and Li, Jiang and Gao, Zijian and others},
  journal={Patterns},
  volume={3},
  number={6},
  year={2022},
  publisher={Elsevier}
}

@article{rasti2017macular,
  title={Macular OCT classification using a multi-scale convolutional neural network ensemble},
  author={Rasti, Reza and Rabbani, Hossein and Mehridehnavi, Alireza and Hajizadeh, Fedra},
  journal={IEEE transactions on medical imaging},
  volume={37},
  number={4},
  pages={1024--1034},
  year={2017},
  publisher={IEEE}
}

@article{abramoff2013automated,
  title={Automated analysis of retinal images for detection of referable diabetic retinopathy},
  author={Abr{\`a}moff, Michael D and Folk, James C and Han, Dennis P and Walker, Jonathan D and Williams, David F and Russell, Stephen R and Massin, Pascale and Cochener, Beatrice and Gain, Philippe and Tang, Li and others},
  journal={JAMA ophthalmology},
  volume={131},
  number={3},
  pages={351--357},
  year={2013},
  publisher={American Medical Association}
}

@article{kermany2018identifying,
  title={Identifying medical diagnoses and treatable diseases by image-based deep learning},
  author={Kermany, Daniel S and Goldbaum, Michael and Cai, Wenjia and Valentim, Carolina CS and Liang, Huiying and Baxter, Sally L and McKeown, Alex and Yang, Ge and Wu, Xiaokang and Yan, Fangbing and others},
  journal={cell},
  volume={172},
  number={5},
  pages={1122--1131},
  year={2018},
  publisher={Elsevier}
}

@inproceedings{hassan2018biomisa,
  title={BIOMISA retinal image database for macular and ocular syndromes},
  author={Hassan, Taimur and Akram, M Usman and Masood, M Furqan and Yasin, Ubaidullah},
  booktitle={International Conference Image Analysis and Recognition},
  pages={695--705},
  year={2018},
  organization={Springer}
}

@article{ma2020rose,
  title={ROSE: a retinal OCT-angiography vessel segmentation dataset and new model},
  author={Ma, Yuhui and Hao, Huaying and Xie, Jianyang and Fu, Huazhu and Zhang, Jiong and Yang, Jianlong and Wang, Zhen and Liu, Jiang and Zheng, Yalin and Zhao, Yitian},
  journal={IEEE transactions on medical imaging},
  volume={40},
  number={3},
  pages={928--939},
  year={2020},
  publisher={IEEE}
}

@article{liao2025unleashing,
  title={Unleashing the potential of open-set noisy samples against label noise for medical image classification},
  author={Liao, Zehui and Hu, Shishuai and Zhang, Yanning and Xia, Yong},
  journal={Medical Image Analysis},
  pages={103702},
  year={2025},
  publisher={Elsevier}
}

@article{rudin2019stop,
  title={Stop explaining black box machine learning models for high stakes decisions and use interpretable models instead},
  author={Rudin, Cynthia},
  journal={Nature machine intelligence},
  volume={1},
  number={5},
  pages={206--215},
  year={2019},
  publisher={Nature Publishing Group UK London}
}

@article{van2022explainable,
  title={Explainable artificial intelligence (XAI) in deep learning-based medical image analysis},
  author={Van der Velden, Bas HM and Kuijf, Hugo J and Gilhuijs, Kenneth GA and Viergever, Max A},
  journal={Medical image analysis},
  volume={79},
  pages={102470},
  year={2022},
  publisher={Elsevier}
}

@article{tang2022fusionm4net,
  title={FusionM4Net: A multi-stage multi-modal learning algorithm for multi-label skin lesion classification},
  author={Tang, Peng and Yan, Xintong and Nan, Yang and Xiang, Shao and Krammer, Sebastian and Lasser, Tobias},
  journal={Medical Image Analysis},
  volume={76},
  pages={102307},
  year={2022},
  publisher={Elsevier}
}

@article{wong2014global,
  title={Global prevalence of age-related macular degeneration and disease burden projection for 2020 and 2040: a systematic review and meta-analysis},
  author={Wong, Wan Ling and Su, Xinyi and Li, Xiang and Cheung, Chui Ming G and Klein, Ronald and Cheng, Ching-Yu and Wong, Tien Yin},
  journal={The Lancet Global Health},
  volume={2},
  number={2},
  pages={e106--e116},
  year={2014},
  publisher={Elsevier}
}

@article{sivaprasad2012prevalence,
  title={Prevalence of diabetic retinopathy in various ethnic groups: a worldwide perspective},
  author={Sivaprasad, Sobha and Gupta, Bhaskar and Crosby-Nwaobi, Roxanne and Evans, Jennifer},
  journal={Survey of ophthalmology},
  volume={57},
  number={4},
  pages={347--370},
  year={2012},
  publisher={Elsevier}
}

@article{thylefors1994global,
  title={The global impact of glaucoma},
  author={Thylefors, B and Negrel, AD2486713},
  journal={Bulletin of the World Health Organization},
  volume={72},
  number={3},
  pages={323},
  year={1994}
}

@article{li2022comparison,
  title={Comparison of fundus fluorescein angiography and fundus photography grading criteria for early diabetic retinopathy},
  author={Li, Xin-Yue and Wang, Shu and Dong, Li and Zhang, Hong},
  journal={International Journal of Ophthalmology},
  volume={15},
  number={2},
  pages={261},
  year={2022}
}

@article{mahendradas2021role,
  title={Role of ocular imaging in diagnosis and determining response to therapeutic interventions in posterior and panuveitis},
  author={Mahendradas, Padmamalini and Sridharan, Akhila and Kawali, Ankush and Sanjay, Srinivasan and Venkatesh, Ramesh},
  journal={The Asia-Pacific Journal of Ophthalmology},
  volume={10},
  number={1},
  pages={74--86},
  year={2021},
  publisher={LWW}
}

@article{invernizzi2023experts,
  title={Experts Opinion: OCTA vs. FFA/ICG in Uveitis--Which Will Survive? “Ten questions to find one answer”},
  author={Invernizzi, Alessandro and Carre{\~n}o, Ester and Pichi, Francesco and Munk, Marion R and Agarwal, Aniruddha and Zierhut, Manfred and Pavesio, Carlos},
  journal={Ocular Immunology and Inflammation},
  volume={31},
  number={8},
  pages={1561--1568},
  year={2023},
  publisher={Taylor \& Francis}
}

@article{jin2022multimodal,
  title={Multimodal deep learning with feature level fusion for identification of choroidal neovascularization activity in age-related macular degeneration},
  author={Jin, Kai and Yan, Yan and Chen, Menglu and Wang, Jun and Pan, Xiangji and Liu, Xindi and Liu, Mushui and Lou, Lixia and Wang, Yao and Ye, Juan},
  journal={Acta Ophthalmologica},
  volume={100},
  number={2},
  pages={e512--e520},
  year={2022},
  publisher={Wiley Online Library}
}

@inproceedings{luo2023harvard,
  title={Harvard glaucoma detection and progression: A multimodal multitask dataset and generalization-reinforced semi-supervised learning},
  author={Luo, Yan and Shi, Min and Tian, Yu and Elze, Tobias and Wang, Mengyu},
  booktitle={Proceedings of the IEEE/CVF International Conference on Computer Vision},
  pages={20471--20482},
  year={2023}
}

@article{hervella2022multimodal,
  title={Multimodal image encoding pre-training for diabetic retinopathy grading},
  author={Hervella, Alvaro S and Rouco, Jos{\'e} and Novo, Jorge and Ortega, Marcos},
  journal={Computers in Biology and Medicine},
  volume={143},
  pages={105302},
  year={2022},
  publisher={Elsevier}
}

@inproceedings{zhou2023representation,
  title={Representation, alignment, fusion: A generic transformer-based framework for multi-modal glaucoma recognition},
  author={Zhou, You and Yang, Gang and Zhou, Yang and Ding, Dayong and Zhao, Jianchun},
  booktitle={International Conference on Medical Image Computing and Computer-Assisted Intervention},
  pages={704--713},
  year={2023},
  organization={Springer}
}

@article{li2023transforming,
  title={Transforming medical imaging with Transformers? A comparative review of key properties, current progresses, and future perspectives},
  author={Li, Jun and Chen, Junyu and Tang, Yucheng and Wang, Ce and Landman, Bennett A and Zhou, S Kevin},
  journal={Medical image analysis},
  volume={85},
  pages={102762},
  year={2023},
  publisher={Elsevier}
}

@article{liu2024multi,
  title={Multi-branch CNN and grouping cascade attention for medical image classification},
  author={Liu, Shiwei and Yue, Wenwen and Guo, Zhiqing and Wang, Liejun},
  journal={Scientific Reports},
  volume={14},
  number={1},
  pages={15013},
  year={2024},
  publisher={Nature Publishing Group UK London}
}

@article{ong2024shortcut,
  title={Shortcut learning in medical AI hinders generalization: method for estimating AI model generalization without external data},
  author={Ong Ly, Cathy and Unnikrishnan, Balagopal and Tadic, Tony and Patel, Tirth and Duhamel, Joe and Kandel, Sonja and Moayedi, Yasbanoo and Brudno, Michael and Hope, Andrew and Ross, Heather and others},
  journal={NPJ digital medicine},
  volume={7},
  number={1},
  pages={124},
  year={2024},
  publisher={Nature Publishing Group UK London}
}

@article{hasani2022trustworthy,
  title={Trustworthy artificial intelligence in medical imaging},
  author={Hasani, Navid and Morris, Michael A and Rhamim, Arman and Summers, Ronald M and Jones, Elizabeth and Siegel, Eliot and Saboury, Babak},
  journal={PET clinics},
  volume={17},
  number={1},
  pages={1},
  year={2022}
}

@inproceedings{guo2017calibration,
  title={On calibration of modern neural networks},
  author={Guo, Chuan and Pleiss, Geoff and Sun, Yu and Weinberger, Kilian Q},
  booktitle={International conference on machine learning},
  pages={1321--1330},
  year={2017},
  organization={PMLR}
}

@article{kendall2017uncertainties,
  title={What uncertainties do we need in bayesian deep learning for computer vision?},
  author={Kendall, Alex and Gal, Yarin},
  journal={Advances in neural information processing systems},
  volume={30},
  year={2017}
}

@article{yang2022machine,
  title={Machine learning generalizability across healthcare settings: insights from multi-site COVID-19 screening},
  author={Yang, Jenny and Soltan, Andrew AS and Clifton, David A},
  journal={NPJ digital medicine},
  volume={5},
  number={1},
  pages={69},
  year={2022},
  publisher={Nature Publishing Group UK London}
}

@inproceedings{yang2021medmnist,
  title={Medmnist classification decathlon: A lightweight automl benchmark for medical image analysis},
  author={Yang, Jiancheng and Shi, Rui and Ni, Bingbing},
  booktitle={2021 IEEE 18th International Symposium on Biomedical Imaging (ISBI)},
  pages={191--195},
  year={2021},
  organization={IEEE}
}

@inproceedings{subramanian2022classification,
  title={Classification of retinal oct images using deep learning},
  author={Subramanian, Malliga and Shanmugavadivel, Kogilavani and Naren, Obuli Sai and Premkumar, K and Rankish, K},
  booktitle={2022 international conference on computer communication and informatics (ICCCI)},
  pages={1--7},
  year={2022},
  organization={IEEE}
}

@article{lin2025healthgpt,
  title={Healthgpt: A medical large vision-language model for unifying comprehension and generation via heterogeneous knowledge adaptation},
  author={Lin, Tianwei and Zhang, Wenqiao and Li, Sijing and Yuan, Yuqian and Yu, Binhe and Li, Haoyuan and He, Wanggui and Jiang, Hao and Li, Mengze and Song, Xiaohui and others},
  journal={arXiv preprint arXiv:2502.09838},
  year={2025}
}

@article{zhang2024hyperllava,
  title={Hyperllava: Dynamic visual and language expert tuning for multimodal large language models},
  author={Zhang, Wenqiao and Lin, Tianwei and Liu, Jiang and Shu, Fangxun and Li, Haoyuan and Zhang, Lei and Wanggui, He and Zhou, Hao and Lv, Zheqi and Jiang, Hao and others},
  journal={arXiv preprint arXiv:2403.13447},
  year={2024}
}

@inproceedings{li2025eyecaregpt,
  title={Eyecaregpt: Boosting comprehensive ophthalmology understanding with tailored dataset, benchmark and model},
  author={Li, Sijing and Lin, Tianwei and Lin, Lingshuai and Zhang, Wenqiao and Liu, Jiang and Yang, Xiaoda and Li, Juncheng and He, Yucheng and Song, Xiaohui and Xiao, Jun and others},
  booktitle={Proceedings of the 33rd ACM International Conference on Multimedia},
  pages={3893--3902},
  year={2025}
}

@article{xie2025heartcare,
  title={Heartcare Suite: Multi-dimensional Understanding of ECG with Raw Multi-lead Signal Modeling},
  author={Xie, Yihan and Li, Sijing and Lin, Tianwei and Wang, Zhuonan and Yang, Chenglin and Zhong, Yu and Zhang, Wenqiao and Li, Haoyuan and Jiang, Hao and Zhang, Fengda and others},
  journal={arXiv preprint arXiv:2506.05831},
  year={2025}
}

@inproceedings{zhang2022boostmis,
  title={Boostmis: Boosting medical image semi-supervised learning with adaptive pseudo labeling and informative active annotation},
  author={Zhang, Wenqiao and Zhu, Lei and Hallinan, James and Zhang, Shengyu and Makmur, Andrew and Cai, Qingpeng and Ooi, Beng Chin},
  booktitle={Proceedings of the IEEE/CVF conference on computer vision and pattern recognition},
  pages={20666--20676},
  year={2022}
}

@inproceedings{lv2023duet,
  title={Duet: A tuning-free device-cloud collaborative parameters generation framework for efficient device model generalization},
  author={Lv, Zheqi and Zhang, Wenqiao and Zhang, Shengyu and Kuang, Kun and Wang, Feng and Wang, Yongwei and Chen, Zhengyu and Shen, Tao and Yang, Hongxia and Ooi, Beng Chin and others},
  booktitle={Proceedings of the ACM Web Conference 2023},
  pages={3077--3085},
  year={2023}
}

@inproceedings{zhang2024revisiting,
  title={Revisiting the domain shift and sample uncertainty in multi-source active domain transfer},
  author={Zhang, Wenqiao and Lv, Zheqi and Zhou, Hao and Liu, Jia-Wei and Li, Juncheng and Li, Mengze and Li, Yunfei and Zhang, Dongping and Zhuang, Yueting and Tang, Siliang},
  booktitle={Proceedings of the IEEE/CVF Conference on Computer Vision and Pattern Recognition},
  pages={16751--16761},
  year={2024}
}

@inproceedings{yuan2025videorefer,
  title={Videorefer suite: Advancing spatial-temporal object understanding with video llm},
  author={Yuan, Yuqian and Zhang, Hang and Li, Wentong and Cheng, Zesen and Zhang, Boqiang and Li, Long and Li, Xin and Zhao, Deli and Zhang, Wenqiao and Zhuang, Yueting and others},
  booktitle={Proceedings of the Computer Vision and Pattern Recognition Conference},
  pages={18970--18980},
  year={2025}
}

@article{yuan2025pixelrefer,
  title={PixelRefer: A Unified Framework for Spatio-Temporal Object Referring with Arbitrary Granularity},
  author={Yuan, Yuqian and Zhang, Wenqiao and Li, Xin and Wang, Shihao and Li, Kehan and Li, Wentong and Xiao, Jun and Zhang, Lei and Ooi, Beng Chin},
  journal={arXiv preprint arXiv:2510.23603},
  year={2025}
}

\end{document}